\documentclass{IEEEcsmag}

\usepackage[colorlinks,urlcolor=blue,linkcolor=blue,citecolor=blue]{hyperref}

\usepackage{upmath}
\usepackage{booktabs}
\usepackage[utf8]{inputenc}
\usepackage{times}
\usepackage{latexsym}
\usepackage{url}
\usepackage{booktabs}
\usepackage{float}
\usepackage{graphicx}
\usepackage{amsmath}

\DeclareMathOperator*{\argmax}{arg\,max}

\newcommand\blfootnote[1]{%
  \begingroup
  \renewcommand\thefootnote{}\footnote{#1}%
  \addtocounter{footnote}{-1}%
  \endgroup
}

\jvol{XX}
\jnum{XX}
\paper{8}
\jmonth{-}
\jname{IEEE Intelligent Systems}
\pubyear{2021}

\begin{document}


\title{Transformer-Based Models for Automatic Identification of Argument Relations: A Cross-Domain Evaluation}

\author{Ramon Ruiz-Dolz}
\affil{Universitat Politècnica de València, Valencian Research Institute for Artificial Intelligence (VRAIN)}

\author{Jose Alemany}
\affil{Universitat Politècnica de València, Valencian Research Institute for Artificial Intelligence (VRAIN)}

\author{Stella Heras}
\affil{Universitat Politècnica de València, Valencian Research Institute for Artificial Intelligence (VRAIN)}

\author{Ana García-Fornes}
\affil{Universitat Politècnica de València, Valencian Research Institute for Artificial Intelligence (VRAIN)}

\markboth{Department Head}{Paper title}

\begin{abstract}
 Argument Mining is defined as the task of automatically identifying and extracting argumentative components (e.g., premises, claims, etc.) and detecting the existing relations among them (i.e., support, attack, rephrase, no relation). One of the main issues when approaching this problem is the lack of data, and the size of the publicly available corpora. In this work, we use the recently annotated \textit{US2016} debate corpus. \textit{US2016} is the largest existing argument annotated corpus, which allows exploring the benefits of the most recent advances in Natural Language Processing in a complex domain like Argument (relation) Mining. We present an exhaustive analysis of the behavior of transformer-based models (i.e., BERT, XLNET, RoBERTa, DistilBERT and ALBERT) when predicting argument relations. Finally, we evaluate the models in five different domains, with the objective of finding the less domain dependent model. We obtain a macro F1-score of 0.70 with the \textit{US2016} evaluation corpus, and a macro F1-score of 0.61 with the \textit{Moral Maze} cross-domain corpus. 
\end{abstract}

\maketitle

\chapterinitial{Computational Argumentation}\blfootnote{\textbf{© 2021 IEEE.  Personal use of this material is permitted.  Permission from IEEE must be obtained for all other uses, in any current or future media, including reprinting/republishing this material for advertising or promotional purposes, creating new collective works, for resale or redistribution to servers or lists, or reuse of any copyrighted component of this work in other works.}} has proved to be a very solid way to approach several problems such as fake news detection \cite{kotonya2019gradual}, recommendation systems \cite{rago2018argumentation} or debate analysis \cite{jha2019formal} among others. However, in almost every domain, it is of great importance to be able to automatically extract the arguments and their relations from the input source. Argument Mining (AM) is the Natural Language Processing (NLP) task by which this problem is addressed. The Transformer model architecture \cite{vaswani2017attention} and its subsequent pre-training approaches have been a turning point in the NLP research area. Thanks to its architecture, it has been possible to capture longer-range dependencies between input structures, and thus the performance of systems developed for the most general NLP tasks (i.e., translation, text generation or language understanding) improved significantly. Therefore, the Transformer architecture has laid the foundations on which newer models and pre-training approaches have been proposed, defining the state-of-the-art in NLP. In this work, we analyze the behavior of BERT \cite{devlin2018bert}, XLNET \cite{yang2019xlnet}, RoBERTa \cite{liu2019roberta}, DistilBERT \cite{sanh2019distilbert} and ALBERT \cite{lan2019albert} when facing the hardest AM task: identifying relational properties between arguments. 

Argument Mining was formally defined in \cite{palau2009argumentation} as the task that aims to automatically detect arguments, relations and their internal structure. As pointed out in \cite{survey}, due to the complexity of AM, the whole task can be decomposed into three main sub-tasks depending on their argumentative complexity. First, the identification of argument components consists in distinguishing argumentative propositions from non-argumentative propositions. This allows to segment the input text into arguments, making it possible to carry out the subsequent sub-tasks. Second, the identification of clausal properties is the part of AM that focuses on finding premises or conclusions among the argumentative propositions. Third, the last sub-task is the identification of relational properties. Two different argumentative propositions are considered at a time, and the main objective is to identify which type of relation links both propositions. Different relations can be observed in argument analysis, from the classical attack/support binary analysis \cite{cocarascu2017identifying}, to the identification of complex patterns of human reasoning (i.e., argumentation schemes \cite{walton2008argumentation}). Therefore, the identification of argumentative relations is the most complex part of AM \cite{survey}, but its complexity may vary depending on how the problem is instantiated.

One of the main problems when addressing any AM task is the lack of high quality annotated data. In fact, as the argumentative complexity of the task increases, it gets harder to find large enough corpus to do experiments that match the latest NLP advances. An important feature that characterizes the transformer-based models is that large corpora are needed to achieve the performance improvement mentioned above. Recently, in \cite{visser2019argumentation}, a new argument annotated corpus of the United States 2016 debate (\textit{US2016}) was published. This corpus contains data from the transcripts of the televised political debates and from internet debates generated around the same context. This is the first publicly available corpus with enough data to begin exploring the benefits of the most recent contributions in NLP, when applied to the identification of argumentative relations. Additionally, the \textit{US2016} corpus has been annotated using Inference Anchoring Theory (IAT\footnote{\url{https://typo.uni-konstanz.de/add-up/wp-content/uploads/2018/04/IAT-CI-Guidelines.pdf}}), a standard argument annotation guideline that provides more information than the classic attack/support binary annotation. Learning a model to automatically annotate with the use of IAT, makes it possible to evaluate its performance not only with the test samples of the corpus, but also with other different corpus already analyzed and tagged using this standard (e.g., \textit{Moral Maze} corpus).
This way, it is our objective to both: evaluate the performance of these new models in the identification of argument relations task; and to find out which one is more robust to variations in the application domain.

In this work, we explore the benefits of the most recent advances in NLP applied to relation prediction in the AM domain. For this purpose we use the recently published \textit{US2016} corpus, since it is to the best of our knowledge, the largest annotated corpus containing information about argumentative relations, and the \textit{Moral Maze} cross-domain corpus. Then we do: (i) a pre-processing of the corpus in order to clean and structure the data for the requirements of our experiments; (ii) an analysis of the performance of the most relevant transformer-based models (i.e., BERT, XLNET, RoBERTa, DistilBERT and ALBERT) when learning to predict the relations between argumentative propositions defined by the IAT standard; and (iii) an evaluation of the obtained models in five different domains (\textit{Moral Maze} corpus) with the objective of analyzing the domain dependency of the transformer-based models when facing this AM task.


\section{RELATED WORK}
\label{rel_work}

Argument Mining is one of the main research areas in Computational Argumentation. AM has caught the attention of many researchers since it is considered to be the first step towards autonomous argumentative systems. We identified many different approaches to the Argument (relation) Mining problem, which depend on the proposed methods (i.e., Parsing algorithms, Textual Entailment Suites, Logistic Regression, Support Vector Machines and Neural Networks), and the available corpus at each moment. Initial research on automatic identification of argument relations was done in \cite{palau2009argumentation} where parsing algorithms were used to determine the type of relation existing between two argument propositions. Some years later, AM started to gain relevance in the NLP community. We can observe the popularization of machine learning techniques for NLP purposes in \cite{naderi2015argumentation}, \cite{stab2017parsing} and \cite{menini2018never}. Support Vector Machines (SVMs) seemed to be the best performing machine learning technique for the purpose of argument relation identification. With the advent of Neural Networks (NNs), a performance gap between previous works and this new approach could be observed. In \cite{cocarascu2017identifying} the empirical results obtained by Recurrent Neural Network (RNN) models for AM were significantly better. However, there is an interesting observation to make emphasis on, which makes it hard to compare AM works. As it can be pointed out after looking at the results depicted in works like \cite{niculae2017argument} or \cite{cocarascu2017identifying}, the corpus used in each work has a lot of influence in the results. This is due to many different factors such as class distributions, variable language complexity (e.g., use of irony, enthymemes, etc.) or the own size of the corpus. Therefore, misleading results may be observed if the generalization of the model is not properly evaluated. On the other hand, deep learning algorithms require much more data to significantly increase its performance compared to classic neural, machine learning or statistical methods. Therefore, from all these past years of argument relation identification works, the performance has been improved not only with the use of new models or techniques, but also with the creation of better corpora. In \cite{visser2019argumentation}, a new argument annotated corpus (\textit{US2016}) was published, with enough data to begin exploiting the benefits of the most recent advances in NLP (i.e., transformer-based models) in the AM domain. To the best of our knowledge, this is the first work addressing the Argument (relation) Mining problem using Transformer-based models in more than a unique domain.

\section{DATA}
\label{corpus}

\begin{figure*}
    \centering
    \includegraphics[width=\textwidth]{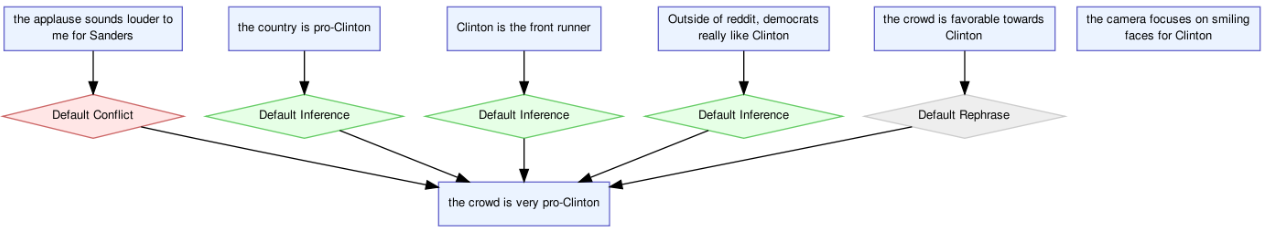}
    \caption{US2016 Argument Map Sample. ADUs are bounded by rectangles. Relation types are contained in the rhombuses.}
    \label{fig:argmap}
\end{figure*}

Two different corpora have been used in this work: the \textit{US2016} debate corpus and the \textit{Moral Maze} multi-domain corpus. Both corpora can be downloaded from the Argument Interchange Format Database (AIFdb), an initiative of researchers from the ARG-tech\footnote{\url{https://www.arg-tech.org/}} with the objective of creating a standard formatted argument corpus database \cite{lawrence2014aifdb}. This database contains 193 different argumentative corpora structured using the AIF standard. Each corpus is divided into several argument maps (\textbf{Figure \ref{fig:argmap}}), and each argument map contains a set of Argument Discourse Units (ADUs) with its argumentative relations annotated using the Inference Anchoring Theory (IAT). This annotation method considers the most important three argumentative relations: inference (RA), conflict (CA) and rephrase (MA). An inference relation between two propositions determines that one is used to support or justify the other; a conflict relation indicates that two propositions have contradictory information; and a rephrase between two propositions means that they are equivalent from an argumentative point of view.

In order to adapt the AIFdb corpus to the needs of this task, we did some pre-processing. Each argument map is stored in a JSON file, and represented as a graph following the AIF standard. We generate a unique tab-separated values file per corpus containing three different values: proposition1, proposition2 and label. 
In addition to the existing IAT relation labels, we decided to generate an additional relation: the no relation (NO) label. Since most of the pairs of propositions found in a debate are not related, we decided to generate a 65\% of samples belonging to this new class. For this purpose, we mixed up the propositions that were not annotated with any of the IAT relation classes. This way, the resulting model will also be able to discriminate between related or not related propositions.

\subsection{\textit{US2016} Debate Corpus}

The \textit{US2016}\footnote{\url{http://corpora.aifdb.org/US2016}} corpus is an argument annotated corpus of the electoral debate carried out in 2016 in the United States. It contains both, transcriptions of the different rounds of TV debate, and discussions from the Reddit forums as detailed in \cite{visser2019argumentation}. The class distribution of the processed \textit{US2016} corpus is depicted in \textbf{Table \ref{tab:us2016data}}. Since it is the largest publicly available argument annotated corpus in the literature, we used it to train the models. We decided to split the corpus with the 80\% of the proposition pairs for training, and the remaining 20\% for the evaluation.

\begin{table}
    \centering
    \caption{Class distribution of the US2016 corpus, Train and Test partitions. }
    \begin{tabular}{l c c c}
    \toprule
         & \textbf{US2016} & \textbf{Train} & \textbf{Test} \\ \midrule
        \textbf{RA} & 2744 & 2195 & 549 \\ \midrule
        \textbf{CA} & 888 & 710 & 178 \\ \midrule
        \textbf{MA} & 705 & 564 & 141 \\ \midrule
        \textbf{NO} & 8055 & 6444 & 1611 \\ \midrule\midrule
        \textbf{Total} & 12392 & 9913 & 2479 \\\bottomrule
    \end{tabular}
    \label{tab:us2016data}
\end{table}

\subsection{\textit{Moral Maze} Multi-Domain Corpus}

The \textit{Moral Maze}\footnote{\url{http://corpora.aifdb.org/mm2012}} multi-domain corpus is an argument annotated corpus obtained from the transcriptions of the 2012 Moral Maze BBC discussion show. This corpus has been built from samples collected in five different broadcasts. The class distribution of the processed \textit{Moral Maze} corpus is depicted in \textbf{Table \ref{tab:mm2012data}}. This corpus is used to evaluate the domain robustness of the trained models across five different domain corpus: Bank (B), Empire (E), Money (M), Problem (P) and Welfare (W); each one focused on a specific debate topic and with a different distribution of classes.

\begin{table}
    \centering
    \caption{Multi-domain evaluation corpus (Moral Maze) class distribution.}
    \resizebox{0.48\textwidth}{!}{%
    \begin{tabular}{l  c c c c c c}
    \toprule
         & \textbf{MM2012} & \textbf{B} & \textbf{E} & \textbf{M} & \textbf{P} & \textbf{W} \\ \midrule
        \textbf{RA} & 833 & 128 & 121  & 205 & 192 & 187 \\ \midrule
        \textbf{CA} & 200 & 26 & 36  & 30 & 45 & 63 \\ \midrule
        \textbf{MA} & 156 & 3 & 25  & 48 & 41 & 39 \\ \midrule
        \textbf{NO} & 2209 & 292 & 339  & 526 & 517 & 537 \\ \midrule\midrule
        \textbf{Total} & 3398 & 449 & 521  & 810 & 795 & 826 \\\bottomrule
    \end{tabular}}
    \label{tab:mm2012data}
\end{table}

\section{AUTOMATIC IDENTIFICATION OF RELATIONAL PROPERTIES}
\label{model}

The problem addressed in this paper can be seen as an instance of the sentence pair classification problem. The sentence pair classification problem consists of assigning the most likely class to two text inputs at a time. In Argument Mining, after segmenting the text and defining the argument components, the argument graph must be built by identifying the relational properties between every two argument components. Therefore, given two argument components (i.e., sentences): $x_1^N = x_1, x_2, \dots, x_N$ of length $N$ where $x_n$ is each word of the first component; and $y_1^M = y_1, y_2, \dots, y_M$ of length $M$ where $y_m$ is each word of the second component, the classification problem can be modeled as defined in Equation \ref{conditional_prob},

\begin{equation}
\label{conditional_prob}
    \hat{c} = \argmax_{c \in C} p(c | x_1^N , y_1^M)
\end{equation}

where $C = [RA, CA, MA, NO]$, so the four different relation types existing in the IAT labeling are considered: inference (RA), conflict (CA), rephrase (MA) and no relation (NO). To approach this problem, we decided to use transformer-based neural architectures. The most recent works in the literature tackle the Argument Mining problem using Recurrent Neural Networks (e.g., LSTMs, BiLSTMs, etc.). However, the Transformer architecture presents several interesting improvements with respect to the RNNs. The Transformer architecture uses multiple attention modules, which allow to capture longer range dependencies between words in a sentence. Given the nature of this work's task, we expect to have long input sentences since argumentative text is, generally, more complex than others. Therefore, we think attention mechanisms can be very useful for the identification of relational properties between argument components.

In this work, we apply Inductive Transfer Learning combined with different Transformer pre-training methods that allow us to learn our task not from scratch but using previously calculated weights. We decided to use the pre-training methods that performed the best in other NLP tasks such as Natural Language Understanding, Question Answering or Text Generation: BERT \cite{devlin2018bert}, XLNET \cite{yang2019xlnet}, RoBERTa \cite{liu2019roberta}, DistilBERT \cite{sanh2019distilbert} and ALBERT \cite{lan2019albert}. All these models have in common that they are based on the Transformer architecture, however different approaches have been considered in order to compute the initial weights. BERT, also known as Bidirectional Encoder Representations from Transformers, is pre-trained on masked language model and next sentence prediction tasks. The model is designed to be able to fine-tune its weights on other different tasks by adding an additional output layer. XLNet is proposed after identifying a potential problem in BERT: the language modeling of the existing dependencies between the masked positions. XLNet combines both auto-regressive language modeling and auto-encoding techniques in order to overcome the detected potential issues. RoBERTa is an strong optimization of the BERT pre-training approach. After researchers did a thorough analysis on the impact of the most important hyper-parameters, this new model was able to obtain interesting results in most of the evaluated tasks. Finally, both DistilBERT and ALBERT were proposed as smaller and faster versions of the previous approaches. We find it interesting to also analyze and evaluate the behavior of these smaller versions, which have been designed to democratize the use of transformer-based pre-training methods without significant loss of performance.

\section{EVALUATION}
\label{eval}

\subsection{Experimental Setup}

All the experiments carried out in this work have been run in a double NVIDIA Titan V computer with an Intel Xeon W-2123 CPU and 62Gb of RAM. This way, we can evaluate not only the performance of the models in the classification task, but also their training computational cost in our specific task. The number of parameters of each model is directly related to the training computational cost. \textbf{Table \ref{tab:modelconfig}} summarises the most relevant features that define each Transformer architecture considered in this research. The Transformer blocks (TBlocks) stand for the number of layers; the hidden size (HSize) represents the number of hidden states in each layer; the attention heads (AH) indicate the number of pointers used by the attention layers; finally, the last feature is the total number of parameters (Params.) of each architecture.


\begin{table}
    \centering
    \caption{Transformer-based architectures configuration.}
    \resizebox{0.48\textwidth}{!}{
    \begin{tabular}{l c c c c}
    \toprule
         \textbf{Model} & \textbf{TBlocks} & \textbf{HSize} & \textbf{AH} & \textbf{Params.} \\ \midrule
        \textbf{BERT-base} \cite{devlin2018bert} & 12 & 768 & 12 & 110M \\ 
        \textbf{BERT-large} \cite{devlin2018bert} & 24 & 1024 & 16 & 340M \\ \midrule
        \textbf{XLNet-base} \cite{yang2019xlnet} & 12 & 768 & 12 & 110M \\ 
        \textbf{XLNet-large} \cite{yang2019xlnet} & 24 & 1024 & 16 & 340M \\ \midrule
        \textbf{RoBERTa-base} \cite{liu2019roberta} & 12 & 768 & 12 & 125M \\ 
        \textbf{RoBERTa-large} \cite{liu2019roberta} & 24 & 1024 & 16 & 335M \\ \midrule
        \textbf{DistilBERT-base} \cite{sanh2019distilbert} & 6 & 768 & 12 & 66M \\ \midrule
        \textbf{ALBERT-base} \cite{lan2019albert} & 12 & 768 & 12 & 11M \\ 
        \textbf{ALBERT-xxlarge} \cite{lan2019albert} & 12 & 4096 & 64 & 223M \\ \bottomrule
    \end{tabular}}
    \label{tab:modelconfig}
\end{table}

In our experiments, we explore the benefits of transfer learning applied to the argument relation mining task. For that purpose, during our training phase, we use the pre-trained encoder of each model with a linear layer on its top. The output size of the linear layer coincides with the number of classes considered in our instance of the problem (i.e., 4). With the \textit{softmax} function, we are able to model the probability of belonging to one class or another for each pair of arguments (Equation \ref{conditional_prob}).

We adapted the maximum sequence length and the batch size of our inputs in each experiment. These parameters were configured in order to use the whole available GPU memory. When training BERT-base models, we defined a maximum sequence length of 256 and a batch size of 64. When training BERT-large models, we halved those values to a maximum sequence length of 128 and a batch size of 32. We trained XLNet-base with a maximum sequence length of 256 and a batch size of 32, and XLNet-large with a maximum sequence length of 256 and a batch size of 8. RoBERTa-base was trained with a maximum sequence length of 256 and batch size of 32, and for training RoBERTa-large we used the same maximum sequence length but a batch size of 16. For DistilBERT we used a maximum sequence length of 256 and a batch size of 128. Finally, ALBERT-base was trained defining a maximum sequence length of 256 and batch size of 64, but in order to fit ALBERT-xxlarge in our available memory we had to define a maximum sequence length of 128 and a batch size of 4. We trained all these models for 50 epochs in our corpus. The best results (depicted in the following section) were obtained with a 1e-5 learning rate.

\subsection{Results}

In this section we present the empirical results obtained after running the experiments on all the previously defined models. In addition to the Transformer-based architectures, we have also trained a Recurrent Neural Network (RNN) as a baseline in our task. We used the best performing RNN architecture in argument relation mining proposed in \cite{cocarascu2017identifying}, consisting of two Long Short-Term Memory (LSTM) networks working in parallel with each pair of arguments. We trained the baseline model for 50 epochs in our data, as the authors did in the original publication. In order to measure the performance of the different models, we have evaluated them using the macro F1-score metric. Due to the huge class imbalance in our corpora, the use of the macro F1-score makes possible to avoid misleading results during the evaluation. Additionally, we also measured the training time required by each model when learning the task proposed in this work, in order to analyze if it can be worthwhile to sacrifice their performance in pursuit of faster training times or availability in lower resource environments.

The macro F1-scores obtained by each model are depicted in \textbf{Table \ref{tab:f1res}}. In the first column, we can see every trained model. The second column represents the macro F1 obtained by each model when evaluated with the test partition of the same corpus used for training (i.e., \textit{US2016}). The third column contains the scores obtained when the evaluation is performed on a different corpus (i.e., \textit{MM2012}) containing a mixture of five domains. Finally, the last five columns are the macro F1-scores of the models when using each one of the five domain specific corpora (i.e., \textit{Bank}, \textit{Empire}, \textit{Money}, \textit{Problem} and \textit{Welfare}) for evaluation.

\begin{table*}
    \centering
    \caption{Performance of the models in the automatic identification of argument relations, given in macro F1-scores.}
    \resizebox{\textwidth}{!}{%
    \begin{tabular}{c c c c c c c c }
    \toprule
         \textbf{Experiment} & \textbf{US2016-test} & \textbf{MM2012} & \textbf{Bank} & \textbf{Empire} & \textbf{Money} & \textbf{Problem} & \textbf{Welfare} \\ \midrule \midrule
        \textbf{LSTM (baseline)} & .26 & .24 & .25 & .22 & .24 & .25 & .23 \\\midrule \midrule
        \textbf{BERT-base-cased} & .62 & .53 & .40 & .45 & .54 & .47 & .53 \\
        \textbf{BERT-base-uncased} & .65 & .56 & .42 & .48 & .54 & .50 & .54 \\
        \textbf{BERT-large-cased} & .61 & .55 & .45 & .49 & .53 & .47 & .51 \\
        \textbf{BERT-large-uncased} & .66 & .57 & .47 & .49 & .56 & .49 & .57 \\\midrule
        \textbf{XLNet-base} & .65 & .56 & .44 & .49 & .51 & .54 & .55 \\
        \textbf{XLNet-large} & .69 & .57 & .44 & .51 & .53 & .53 & .54 \\\midrule
        \textbf{RoBERTa-base} & .68 & .58 & .51 & .52 & .54 & .52 & .58 \\
        \textbf{RoBERTa-large} & \textbf{.70} & \textbf{.61} & \textbf{.53} & .53 & \textbf{.59} & \textbf{.56} & \textbf{.59} \\\midrule
        \textbf{DistilBERT} & .55 & .42 & .33 & .39 & .40 & .43 & .39 \\\midrule
        \textbf{ALBERT-base-v2} & .60 & .54 & .49 & .45 & .53 & .47 & .51 \\
        \textbf{ALBERT-xxlarge-v2} & .67 & .59 & .50 & \textbf{.54} & .56 & .48 & \textbf{.59}\\\bottomrule
    \end{tabular}
    }
    \label{tab:f1res}
\end{table*}

With most of the models, we achieved state-of-the-art macro F1-scores for relation identification in Argument Mining \cite{cocarascu2020dataset}. Here, it is important to make emphasis that the way we considered to represent argumentative relations (i.e., IAT labelling) make this task harder than most of the previous work (i.e., attack/support) in this area. We obtained a 0.70 macro F1-score with \textit{RoBERTa-large}, outperforming the LSTM baseline used as a reference of previous research in argument relation identification. Furthermore, in order to have a more strong reference to compare with previous published results, we carried out an experiment using the same parameters but considering a binary instance of the problem (i.e., only attack and support relations). This way, \textit{RoBERTa-large} achieved a macro F1-score of 0.81 highlighting the mentioned complexity gap between the two instances of the same problem. In general, we can observe that \textit{RoBERTa} has performed very well in this task. When looking at the cross-domain evaluation, \textit{RoBERTa-large} has also performed the best. We obtained a 0.61 macro F1-score when doing the evaluation with a different domain corpus. Moreover, the model has been able to keep a good performance with each one of the five domain specific corpora, even having different class distributions. With \textit{ALBERT-xxlarge-v2} it has been possible to obtain a slightly better performance when evaluating with the \textit{Empire} corpus. It is possible to observe how the scores obtained on the \textit{Bank} and \textit{Empire} corpus are slightly lower than the rest. This is mainly due to their smaller size, combined with the strong imbalance between classes. We also did experiments with cased and uncased models, in order to see the relevance of cased text in the relation identification task. As we can observe, the uncased models performed significantly better than the cased models, so we can point out that cased text did not help to improve the performance of the models in our task.

On the other hand, we obtained the worst results with \textit{DistilBERT} and \textit{ALBERT-base-v2}, as one might expect. We decided to use these models in order to see if the observed performance sacrifice was worthwhile in exchange for more feasible training times. \textbf{Table \ref{tab:times}} contains the training times required by each model under our experimental setup. With \textit{DistilBERT}, it was possible to achieve a significant reduction of training time in exchange for a huge drop in performance. However, with \textit{ALBERT-base-v2} we could not observe a significant reduction of training time. From our experiments, we have not seen any significant advantage in using these \textit{lite} models. We also observed that the computational cost of training \textit{XLNet-large} and \textit{ALBERT-xxlarge-v2} in our task was very expensive. \textit{XLNet-large} was 5.1 times slower to train than \textit{BERT-large}. As for \textit{ALBERT-xxlarge-v2}, the training time was 7.1 times higher than \textit{BERT-large}. This is due to its hidden size of 4096 with respect to the 1024 sized large models. Thus, observing the performance of the models in means of their macro F1-score and the required training time, we still think that \textit{RoBERTa} is the best approach to tackle both, domain-specific and cross-domain identification of relational properties between arguments. Even the \textit{RoBERTa-base} version performed well in this task and it was 6.6 times faster than its large version on training.

\begin{table}
    \centering
    \caption{Training time of 50 epochs running in a double NVIDIA Titan V computer.}
    \begin{tabular}{c  r}
    \toprule
         \textbf{Experiment} & \textbf{Time} \\ \midrule
        \textbf{BERT-base} & 39m 11s \\
        \textbf{BERT-large} & 2h 19m 57s \\\midrule
        \textbf{XLNet-base} & 1h 52m 38s \\
        \textbf{XLNet-large} & 11h 51m 09s \\\midrule
        \textbf{RoBERTa-base} & 43m 17s \\
        \textbf{RoBERTa-large} & 4h 44m 33s \\\midrule
        \textbf{DistilBERT} & 16m 15s \\\midrule
        \textbf{ALBERT-base-v2} & 38m 04s \\
        \textbf{ALBERT-xxlarge-v2} & 16h 20m 22s \\\bottomrule
    \end{tabular}
    \label{tab:times}
\end{table}

\subsection{Error Analysis}

\begin{table}
    \centering
    \caption{Distribution of the misclassified samples per class using the \textit{RoBERTa-large} model. Each column indicates the real class of the samples, each row indicates the assigned class by our model.}
    \begin{tabular}{l c c c c}
    \toprule
         \textbf{Pred. $\backslash$ Real} & \textbf{RA} &\textbf{CA} & \textbf{MA} & \textbf{NO} \\ \midrule
        \textbf{RA} & - & 0.512 & \textbf{0.603} & \textbf{0.730} \\\midrule
        \textbf{CA} & 0.200 & - & 0.138 & 0.226 \\\midrule
        \textbf{MA} & 0.100 & 0.075 & - & 0.044 \\\midrule
        \textbf{NO} & \textbf{0.700} & 0.412 & 0.259 & - \\\bottomrule
    \end{tabular}
    \label{tab:errors}
\end{table}

In an effort to conduct a thorough evaluation, we decided to analyze the errors made by \textit{RoBERTA-large}, the best performing model. For this purpose we measured the volume of misclassifications found on each one of the four classes considered in this work. \textbf{Table \ref{tab:errors}} shows the error distribution detected when analyzing the results. Two important remarks can be pointed out when looking at the obtained error distributions. First of all, it is possible to observe how most of the misclassified argument pairs labeled with an inference relation were assigned the no relation class. Similarly, most of the misclassified argument pairs without relation were assigned the inference class by our model. We observed that many of these errors were due to a loss of contextual information. In an argumentative discourse, it is very common to refer to past concepts without explicitly mentioning them (i.e., enthymemes) or simplifying them with the use of pronouns. The lack of dialogical context can make the automatic identification of argument relations a harder task. For a better understanding of this problem we present the following example with two argument components labeled with inference relation:

\begin{itemize}
    \item[P1:] \textit{I think \textbf{it}’s not going to help change the culture} 
    \item[P2:] \textit{In banking \textbf{we}’ve a totally different situation}
\end{itemize}

Our system classified the pair as no related samples. In fact, by only reading the sentence pair, one may think there is not any argument relation between them. In these situations, it is evident that the key to avoid any possible error is to give additional information about the uttered propositions. In this case, depending on the background meaning of the ``\textit{\textbf{it}}" and ``\textbf{\textit{we}}" pronouns, the sentences may be related or not. The only way of considering this proposition pair related as an inference, is assuming that the \textit{\textbf{it}} pronoun refers to the banking system. We also detected that in these situations the softmax outputs of our model gives very close probabilities for both, \textit{RA} and \textit{NO} classes. Another indicator of the existing model misunderstandings presented before, are the similar error distributions that conflicting arguments show with both inference and no relation classes. On the other hand, we also pointed out that the rephrased argument pairs were mainly misclassified as inference related arguments. However, when analyzing them we observed that most of the relations could also be considered as inference related arguments depending on the interpretation. For example:

\begin{itemize}
    \item[P1:] \textit{\textbf{We} do need curriculum reform}
    \item[P2:] \textit{\textbf{RUBIO too} believes in curriculum reform}
\end{itemize}

In this case, the sentence pair can be interpreted as a rephrase, assuming that ``\textit{\textbf{We}}" and ``\textit{\textbf{RUBIO too}}" are equivalent subjects. But it can also be interpreted as an argument from authority, with P2 supporting (inference relation) P1. In some situations the line that differences rephrase from inference may not be as clear as desired, and both types of relation can be considered correct. Additionally, with these second type of significant detected errors, it is also possible to observe the problem mentioned before. Therefore, the loss of information caused by the use of pronouns or enthymemes in the discourse can be determinant when approaching a task of this complexity.

\section{CONCLUSION}
\label{conclusions}

The automatic identification of argument relations is an essential task in the whole computational argumentation process. It allows to automatically generate the argumentative structure from argument discourse units. In this work, we present how the automatic identification of argument relations, based on Inference Anchoring Theory labeling, can be approached using the latest advances in natural language processing. To the best of our knowledge, this is the first work using transformer-based pre-trained models to learn this task. For this purpose, we have used the largest publicly available argument annotated corpus to the date. Most of the trained models have been able to outperform the state-of-the-art baselines in argument relation mining (\cite{cocarascu2020dataset}), even with a more complex instance of the task. We observed a significant better performance with \textit{RoBERTa} than other models, the best results were achieved with \textit{RoBERTa-large}. We also made a cross-domain evaluation of the models, in order to find out their domain robustness. Even there was a small drop in performance (most probably because of the significant variations of linguistic and class distributions between different domain corpora), the scores on different domains were still close to previous AM reports on a unique domain. This way, it is our objective to contribute on paving the way for finding models that do a better generalisation of this task. Finally, we analyzed the errors made by our best performing model. We have seen that two important groups of errors are caused by the loss of contextual information. We also pointed out that another important group of errors made by the model was due to possible multiple interpretations of the relations. We think that significant improvements in model performance can be achieved after analyzing the most common errors detected in this work. As future work, we propose the following modifications to the automatic identification of argument relations task: (i) pronoun replacement, to solve the loss of contextual information in some propositions; (ii) consider the possible classification ambiguity, in some cases, by accepting multiple correct relations if the interpretation leads to this conclusion; and (iii) incorporation of external information. In argumentation theory, an enthymeme is known as the omission of a claim or a support of an argument. In order to make the discourse more fluid, it is very common to use enthymemes in situations where the omitted information is considered to be known by all the participants. Therefore, without external information, the model may not be able to fully understand relations between enthymemes.

\section{ACKNOWLEDGMENT}

This work is partially supported by the Spanish Government project TIN2017-89156-R, the FPI grant BES-2015-074498, and the Valencian Government project PROMETEO/2018/002. We gratefully acknowledge the support of NVIDIA Corporation with the donation of the Titan V GPUs used for this research.

\bibliographystyle{plain}
\bibliography{paper}

\begin{IEEEbiography}{Ramon Ruiz-Dolz}{\,} received the Master's Degree in Artificial Intelligence, Pattern Recognition and Digital Imaging from the Universitat Politècnica de València (UPV) in 2019. He currently works as a researcher at the Valencian Research Institute for Artificial Intelligence (VRAIN) and is pursuing a Ph.D. degree in Computer Science. His research interests are focused on Argument Mining, Computational Argumentation and Persuasion Technologies. Contact him at the Universitat Politècnica de València, 46022, Valencia, Spain, raruidol@dsic.upv.es.
\end{IEEEbiography}

\begin{IEEEbiography}{Jose Alemany }{\,} received his PhD degree in Computer Science from the Universitat Politècnica de València (UPV) in 2020. He is currently working as a Postdoctoral Researcher at the Valencian Research Institute for Artificial Intelligence (VRAIN). He is also working as an Assistant Professor at Florida Universitària. His research interests include information dissemination, privacy-preserving, content analysis, and complex networks. Contact him at the Universitat Politècnica de València, 46022, Valencia, Spain, jalemany1@dsic.upv.es.
\end{IEEEbiography}

\begin{IEEEbiography}{Stella M. Heras Barberá}{\,} holds a PhD in Computer Science (extraordinary prize Cum Laude) from the Polytechnic University of Valencia (UPV, 2011). She obtained the Executive Master in Project Management by the University of Valencia (2013) (certified PMP-1558995) and has the title of University Specialist in University Pedagogy (UPV, 2011). She currently works as a researcher at the Valencian Research Institute for Artificial Intelligence (VRAIN) and as an associate professor in the Department of Languages and Computer Systems at the UPV, where she has been teaching since 2007. Her research area is focused on the development of artificial intelligence systems (computational argumentation, persuasion technologies, educational recommender systems). Contact her at the Universitat Politècnica de València, 46022, Valencia, Spain, stehebar@upv.es.
\end{IEEEbiography}

\begin{IEEEbiography}{Ana García-Fornes}{\,} received  her  PhD degree  in  Computer  Science from  the Universitat Politècnica de València (UPV, 1996). She currently works as a Full Professor at the Department of Information Systems and Computation at the UPV, Spain and as a researcher at the Valencian Research Institute for Artificial Intelligence (VRAIN).  Her  research  interests  focus  on  real-time  systems, multi-agent systems, agreement technologies and privacy in social media. Contact her at the Universitat Politècnica de València, 46022, Valencia, Spain, agarcia@dsic.upv.es.
\end{IEEEbiography}

\end{document}